\documentclass[letterpaper, 10 pt, conference]{ieeeconf}  

\IEEEoverridecommandlockouts                              

\usepackage{graphics} 
\usepackage{epsfig} 
\usepackage{mathptmx} 
\usepackage{times} 
\usepackage{amsmath} 
\usepackage{amssymb}  
\usepackage{multirow}
\usepackage{booktabs}
\usepackage{ulem} 
\usepackage{cite}
\usepackage{stfloats}
\usepackage{xcolor}
\usepackage[caption=false]{subfig}
\usepackage{hyperref}
\hypersetup{hidelinks,
	colorlinks=true,
	allcolors=black,
	pdfstartview=Fit,
	breaklinks=true}
\newcommand{\warning}[1]{\textcolor[RGB]{0, 0, 0}{#1}}

\title{\LARGE \bf
THE-SEAN: A Heart Rate Variation-Inspired Temporally High-Order Event-Based Visual Odometry with Self-Supervised \\ Spiking Event Accumulation Networks
}

\author{Chaoran Xiong$^1$,~\IEEEmembership{Student~Member,~IEEE}, Litao Wei$^1$, Kehui Ma$^1$, Zhen Sun$^1$, Yan Xiang$^1$, Zihan Nan$^2$, \\ Trieu-Kien Truong$^1$,~\IEEEmembership{Life~Fellow,~IEEE} and Ling Pei$^{1,\ast}$, ~\IEEEmembership{Senior~Member,~IEEE}
\thanks{This work was supported in part by the Basic Science Center Program of the National Natural Science Foundation of China (Grant No.62388101), in part by National Nature Science Foundation of China (NSFC) (Grant Number: 62273229) and in part by Science and Technology Commission of Shanghai Municipality (Grant Number: 24TS1402600 and 24TS1402800).}
\thanks{$^{\ast}$Corresponding author: Ling Pei.}
\thanks{$^1$The authors are with the Shanghai Jiao Tong University, Shanghai 200240,
China (e-mail: sjtu4742986; oscar0371; khma0929; zhensun; yan.xiang; truong@isu.edu.tw; ling.pei@sjtu.edu.cn).}%
\thanks{$^2$Zihan Nan is with Beijing Institute of Aerospace Control Devices, Beijing Institute of Aerospace Control Devices, 100039, Beijing, China (e-mail: nan657584155@163.com).}%
\thanks{The code will be released at \href{https://github.com/Franky-X/THE-SEAN}{https://github.com/Franky-X/THE-SEAN}.}
}

\begin{document}

\maketitle
\thispagestyle{empty}
\pagestyle{empty}

\begin{abstract}
Event-based visual odometry has recently gained attention for its high accuracy and real-time performance in fast-motion systems. Unlike traditional synchronous estimators that rely on constant-frequency (zero-order) triggers, event-based visual odometry can actively accumulate information to generate temporally high-order estimation triggers. However, existing methods primarily focus on adaptive event representation after estimation triggers, neglecting the decision-making process for efficient temporal triggering itself. This oversight leads to the computational redundancy and noise accumulation.
In this paper, we introduce a temporally high-order event-based visual odometry with spiking event accumulation networks (THE-SEAN). To the best of our knowledge, it is the first event-based visual odometry capable of dynamically adjusting its estimation trigger decision in response to motion and environmental changes. Inspired by biological systems that regulate hormone secretion to modulate heart rate, a self-supervised spiking neural network is designed to generate estimation triggers. This spiking network extracts temporal features to produce triggers, with rewards based on block matching points and Fisher information matrix (FIM) trace acquired from the estimator itself.
Finally, THE-SEAN is evaluated across several open datasets, thereby demonstrating average improvements of 13\% in estimation accuracy, 9\% in smoothness, and 38\% in triggering efficiency compared to the state-of-the-art methods.
\end{abstract}

\section{INTRODUCTION}

Event-based visual odometry is a state estimation framework that uses data from event cameras, which offers advantages of low latency, high accuracy, and low power consumption\cite{ESVO, ESVIO_AA, ES-PTAM, ESVO2}. Unlike traditional frame-based synchronous estimators\cite{VINS-Mono, VINS-Sync, ORBSLAM3, 360VIO, TON-VIO} that rely on constant-frequency external triggers (zero-order sampling), event-based estimators autonomously generate internal triggers for estimation\cite{E_Survey}. This capability allows them to adapt dynamically to the motion and environmental context, thus improving performance in fast-motion systems and enhancing information processing efficiency.

\warning{Typically,} event-based estimators require the accumulation of asynchronous event data to process\cite{ESVO, ESVIO_AA, ES-PTAM, ESVO2}. \warning{Various methods have been explored for the event stream representation. Common approaches for event frame representation include constant window methods, such as the naive direct accumulation approach\cite{EAS} and time surface\cite{TS}, which are straightforward to implement and yield intuitive results.} Alternatively, adaptive window methods, such as adaptive time surface\cite{ATS_BM, T-ESVO} and adaptive accumulation\cite{AA}, offer more efficient event representations, \warning{but} they often involve higher computational complexity during the event frame construction at each estimation trigger. \warning{They aim}
to accumulate sufficient information at the estimation trigger while minimizing the accumulation of irrelevant data.

\begin{figure}[t]
    \centering
    \includegraphics[width=1\linewidth]{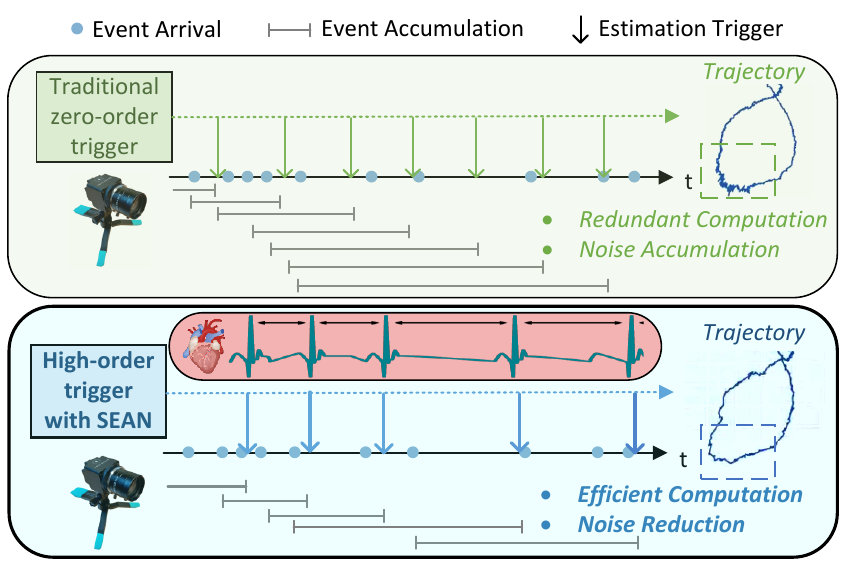}
    \caption{Comparison of the proposed temporally high-order event-based visual odometry system with the traditional zero-order event-based estimator. Traditional estimators typically rely on constant-frequency triggers with adaptive accumulation, resulting in computational redundancy and noise accumulation. In contrast, our proposed temporally high-order system dynamically determines the optimal trigger moments, \warning{which is inspired by heart rate variation mechanism of human}, thereby enhancing both computational efficiency and estimation accuracy.}
    \label{fig:abstract}
    \vspace{-0.5cm}
\end{figure}
However, existing research has primarily focused on event accumulation after the estimation trigger\cite{TS,ATS_BM,T-ESVO,AA}, overlooking the significance of the trigger decision itself. On the one hand, high estimation trigger frequency can lead to redundant computational overhead, especially when using adaptive representations. For instance, \warning{frequent triggering still accumulates events from a longer old time winder under sparse data conditions, resulting in} low information gain and inefficient use of computational resources. On the other hand, low trigger frequency may lead to insufficient information utilization, \warning{degrading} the estimation accuracy, see \cite{EAS}. Therefore, adaptive trigger decisions in the temporal dimension is crucial for event-based estimators. This capability \warning{enables} the estimator to increase the trigger frequency during fast-motion conditions to enhance accuracy, while reducing it in static or low-motion scenarios to \warning{conserve} computational resources and minimize accumulated noise.

In this paper, we propose a temporally high-order event-based (THE) odometry system, \warning{in which} the trigger decision in the temporal domain is generated by spiking event accumulation networks (SEAN). Inspired by biological mechanisms, which regulate hormone and pheromone signaling to adapt to dynamic motion and environmental conditions\cite{Bio}, SEAN is designed to emulate this process. Firstly, SEAN \warning{employs} a leaky integrate-and-fire (LIF) neural network to extract features from the incoming event stream. Then, \warning{the} leaky integrate (LI) neurons of SEAN are used to perform value regression in order to generate rewards for triggering or maintaining idle states. Finally, the weights of SEAN are adjusted based on the information gain derived from the estimator itself. \warning{This} process simulates the biological process of regulating heart rate. To the best of our knowledge, THE-SEAN is the first event-based visual odometry system capable of dynamically adjusting its estimation trigger decision based on its own motion and surrounding environmental changes, \warning{as illustrated in Fig. \ref{fig:abstract}.} \warning{Our main contributions are as follows}:
\begin{enumerate}
    \item A bio-inspired temporally high-order event-based (THE) visual odometry with spiking event accumulation networks (SEAN). \warning{THE-SEAN can dynamically adjusting its mapping and tracking trigger decision policy} based on its motion and surrounding environment, \warning{mimicking} biological mechanisms regulating hormone secretion to modulate heart rate.
    \item A self-supervised Q-learning strategy using the information gain computed from the estimator itself, \warning{including valid block matching points for when to map and Fisher information matrix (FIM) trace for when to track}, as rewards instead of labeled ground truth trajectory. \warning{This ensures that} THE-SEAN can operate effectively across various scenarios without the need for pre-training or a large number of parameters.
    \item New evaluation metrics \warning{including tracking and mapping triggering rate} to assess estimation triggering efficiency for event-based visual odometry. Experiments \warning{conducted} across various open datasets demonstrate that, on average, THE-SEAN improves estimation accuracy by 13\%, enhances smoothness by 8\%, and reduces the estimation trigger rate by 38\% compared to the latest event-based odometry algorithms.
\end{enumerate}

The remainder of this paper is organized as follows: The related work on event-based visual odometry is reviewed in Section II. Section III describes the formulation of decision-making process for the temporally high-order event-based estimator. Section IV introduces the main technical contributions of this work. Section V presents a comparison of THE-SEAN with the latest event-based visual odometry methods across multiple open datasets, including an ablation study. Finally, the conclusion is given in Section VI .

\section{RELATED WORK}
Event-based visual odometry has gained significant attention due to its low latency, low power consumption, and high accuracy in fast-motion systems \cite{ESVO, ESVIO_AA, ES-PTAM, ESVO2}. Unlike traditional cameras that rely on external constant-frequency triggers, event cameras capture asynchronous data, mimicking human vision. Stereo event cameras, in particular, enable more effective depth perception by emulating human binocular vision.

Different from traditional camera-based estimation methods, event-based visual odometry depends on internal mechanisms for accumulating event data, as it lacks external triggers\cite{EAS}. Existing approaches focus on accumulating events after a trigger, typically within a temporal window, and can be categorized into constant-window and adaptive-window accumulation methods. The constant window accumulation methods, such as simple accumulation \cite{EAS} and time surface decay \cite{TS}, are easy to implement but may result in either insufficient or excessive data accumulation. In contrast, adaptive window methods adjust the time window based on the amount of accumulated event information \cite{AA}, ensuring adequate data is available at each trigger. Although more efficient, adaptive representation methods tend to incur higher computational costs and are subject to noise accumulation. 

Building on event representation, various odometry algorithms have been developed. \warning{In \cite{ESVO}, Zhou et al. first proposed to perform mapping and tracking process simultaneously in classic stereo event-based visual odometry (ESVO).} The mapping module estimates depth through stereo disparity, generating reference frames to perform pose estimation in the tracking module. The tracking module projects the accumulated event frames onto these reference frames for pose estimation. \warning{Building on these two modules, different event representation methods have been introduced in the event-based visual odometry, see \cite{ES-PTAM, ESVIO_AA, ESVO2}.} \warning{However}, both mapping and tracking modules of these systems depend on fixed-rate triggers determined by the platform's processing capacity, resulting in considerable computational overhead. Moreover, fixed-frequency triggering can lead to limited information gain in certain situations, \warning{especially when} overwhelmed by noise, negatively impacting estimation accuracy and stability.

To date, existing stereo event-based visual odometry methods often neglect the critical importance of the trigger decision. In constant window methods, optimal trigger decision-making ensures adequate data accumulation at each trigger point. \warning{Similarly,} adaptive methods rely on well-timed trigger decisions to prevent redundant accumulation, especially when event data is sparse. Therefore, \warning{it is essential to adjust the trigger frequency adaptively based on motion dynamics and environmental changes.} Nevertheless, current methods \warning{lack} temporally high-order triggering mechanisms that can adjust based on scene dynamics.

\section{PROBLEM FORMULATION}

This section formulates temporally high-order state estimation for asynchronous event-based systems. Section III-A outlines the key challenges in current asynchronous estimation paradigms. Section III-B introduces a Markov Decision Process (MDP)-based approach to tackle the high-order triggering problem. This formulation enables time-aware triggering policies that optimize the timing of state estimate updates.

\subsection{Asynchronous vs. Synchronous Estimation}

Traditional state estimation systems rely on external triggers to update the state within synchronous frameworks. Conversely, asynchronous event-based estimation autonomously determines state updates based on irregular, event-driven inputs. This introduces challenges in balancing computational efficiency, accuracy, and responsiveness to sparse, temporally irregular events. \warning{However, current event-based estimation methods still rely on constant-frequency triggering, which yields} $\warning{{\operatorname{d} \Delta t} / {\operatorname{d}t} = 0,}$
\warning{where the trigger time interval $\Delta t$ has a derivative equal to zero with respect to time $t$, and thus is referred to as the zero-order estimation.} This synchronous approach is inefficient, resulting in suboptimal resource utilization and poor temporal fusion of information. Existing methods focus on event processing at each trigger, but neglect the crucial task of modeling the temporal process for active asynchronous triggering. Therefore, a new formulation for temporal modeling in asynchronous estimation is needed.

\begin{figure}
    \centering
    \includegraphics[width=0.9\linewidth]{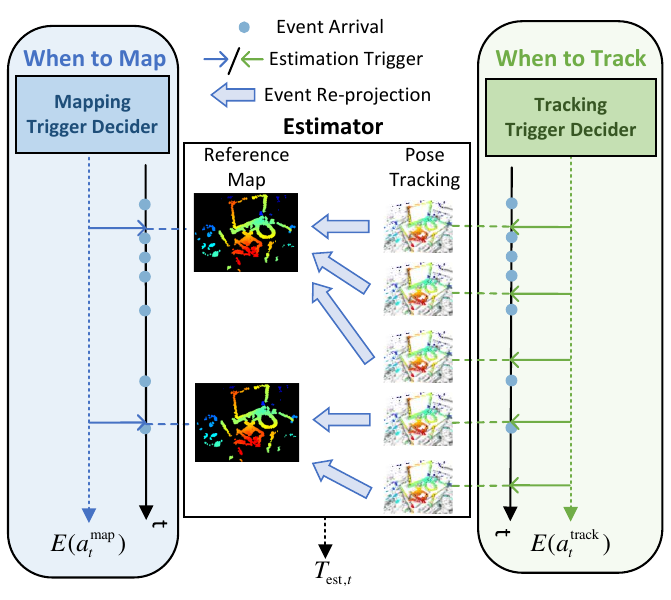}
    \caption{Problem formulation of temporally high-order event-based estimator. The asynchronous estimator must determine when to trigger the mapping and tracking process in order to minimize both estimation error and power consumption.}
    \label{fig:formulation}
    \vspace{-0.5cm}
\end{figure}

\subsection{Formulation of Temporally High-Order Estimation}
To address the lack of principled temporal modeling in existing event-based estimation frameworks, this paper introduces the concept of temporally high-order state estimation shown in Fig. \ref{fig:formulation}. We define asynchronous estimation as a decision-making process where the estimator must determine:

\begin{itemize} 
\item \textbf{When to map}: At each event trigger, the estimator decides whether to create or update the reference depth frame through the mapping process. 
\item \textbf{When to track}: At each event trigger, the estimator determines whether to update the agent’s pose by performing the tracking process. 
\end{itemize}

The decision-making process for when to trigger tracking and mapping in an estimator is modeled as a MDP. The goal is to minimize both estimation error and power consumption. The components of the MDP are outlined below.

\subsubsection{State Representation}

The state \( \mathbf{s}_t \) of the agent at time \( t \) consists of the current input event stream within a temporal window, denoted by
\begin{equation}
    \mathbf{s}_t = \{(x, y, p, i) \mid i \in [t - t_w, t] \},
\end{equation}
where \( x, y \) are the coordinates of active pixels. \( p \) is the polarity of the active event, and \( i \) is the timestamp of the event. \( t_w \) represents the temporal window over which the state is considered for decision-making.

\subsubsection{Action Space}

The action \( \mathbf{a}_t \) represents the decisions the estimator can take, which includes:
\begin{equation}
    \mathbf{a}_t = \{a^{\text{map}}_t,a^{\text{track}}_t\},
\end{equation}
where $a^{\text{map}}_t$ and $a^{\text{track}}_t$ are binary indicators (either 0 or 1) denoting whether the tracking or mapping process should be triggered.

\subsubsection{Energy Consumption}

Each action incurs an associated energy cost for computation. The energy consumption \( E(\mathbf{a}_t) \) for a given action \( \mathbf{a}_t \) is given by
\begin{equation}
   E(\mathbf{a}_t) = a^{\text{map}}_t E_{\text{map}} + a^{\text{track}}_t E_{\text{track}}, 
\end{equation}
where $ E_{\text{map}} $ and  $E_{\text{track}}$ represent the power consumption for the tracking and mapping computations, respectively.

\subsubsection{Policy}

The policy \( \pi(\cdot) \) governs the action selection process and aims to minimize both the estimation error and power consumption. The optimal policy \( \pi^* \) minimizes the following objective function
\begin{equation}
    \pi^* = \arg\min_{\pi} \left(  \lambda_E \frac{1}{N} \sum_{t=0}^{N} E(\mathbf{a}_t) + \lambda_P \frac{1}{N} \sum_{t=1}^{N} \left(T_{g,i}^{-1} * T_{e\text{st},i}\right) \right),
\end{equation}
where \( T_{g,i} \) is the ground truth pose at time step \( i \), and \( T_{e\text{st},i} \) is the estimated pose at the same time. \( \lambda_E \) and \( \lambda_P \) are predefined weights that reflect the relative importance of energy consumption and pose accuracy, respectively. \warning{\( \pi^* \) determines when to map or track for the estimator. Hence, the trigger time interval $\Delta t$ is adaptive with respect to time, and thus is referred to as the high-order estimation.}

\begin{figure*}
    \centering
    \includegraphics[width=1\linewidth]{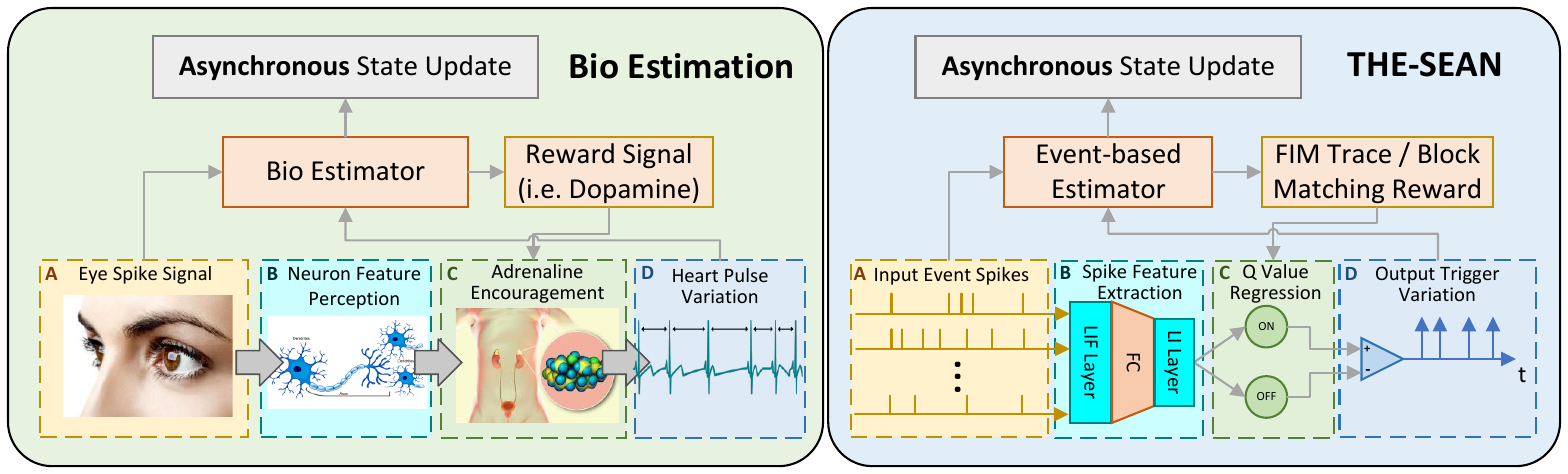}
    \caption{System overview of the bio-inspired temporally high-order estimation framework, THE-SEAN. The left green section illustrates the biological mechanism for asynchronous state estimation, where sensory spike signals pass through neurons, triggering hormone secretion that regulates heart rate. Dopamine, as a reward signal, adjusts hormone levels in a feedback loop. The right blue section shows the proposed THE-SEAN, which emulates this process. Event spikes are processed by spiking neural networks, and Q-values are regressed to regulate the trigger rate. \warning{The rewards acquired from the estimator itself, including Fisher information matrix (FIM) trace and valid block matching points, supervise the network weights in a closed-loop system.} The corresponding processes are color-coded by A, B, C, D for clarity.}
    \label{fig:framework}
    \vspace{-0.5cm}
\end{figure*}
\section{METHODOLOGY}
In this section, we first introduce THE-SEAN, a bio-inspired framework for asynchronous state estimation. We then present the use of spiking neural networks and self supervised reinforcement learning methods for the mapping and tracking process. Finally, the implementation settings of SEAN in event-based estimator are provided.

\subsection{System Overview}

THE-SEAN, our bio-inspired temporally high-order estimation framework is illustrated in Fig. \ref{fig:framework}. As for biological mechanisms, the human eye generates sensory pulses and transmits the signal to the neurons. Then these neurons trigger hormone secretion that regulates heart rate. 
In reality, rapid environmental changes increase hormone secretion by reward signals like dopamine, thereby speeding up the heart rate and enhancing the body’s sensory processing. Conversely, slow environmental changes decrease hormone release, slowing the heart rate and reducing the frequency and sensitivity of state estimation.

Inspired by this biological asynchronous estimation system, we design THE-SEAN, an asynchronous estimation framework for event cameras. Firstly, the pulse signals from the camera are processed by the leaky integrate-and-fire (LIF) neurons to simulate neuron activity. Then, leaky integrate (LI) neurons are used to generate voltage values, which are regressed to ON and OFF values. These values are compared to generate triggers for state updates. Furthermore, network weights are adjusted through self-supervised reward-based learning. Finally, SEAN outputs asynchronous triggers for the event-based estimator and the estimator produces rewards for the network weights adjustment.

\subsection{Architecture and Dynamics of SEAN}
To mimic the function of neuron feature perception for human, an asynchronous spiking event accumulation network (SEAN) is designed to extract the features of the asynchronous event stream. The temporal dynamics of SEAN is introduced as follows.

Firstly, the input event stream is processed through the fully connected leaky integrate-and-fire (LIF) layer of the SNN. The LIF neuron dynamics is described as
\begin{align}
H^i_t & =f\left(V^i_{t-1}, E^i_t\right), \\
S^{i}_t & =\Theta\left(H^i_t-V^i_{th}\right), \\
V^i_t & = H^i_t\left(1-S^{i}_t\right)+V^i_{\text {r}} S^{i}_t,
\end{align}
where $H^i_t$ and $V^i_t$ denote the $i$th membrane voltage after neural dynamics and the trigger of a spike at time-step $t$, respectively. $E^i_t$ denotes the $i$th pixel event trigger input, and $S^{\text{i}}_t$ means the $i$th LIF neuron output spike at time-step $t$, which equals 1 if there is a spike and 0 otherwise. $V_{th}$ denotes the threshold voltage and $V_{\text {r }}$ denotes the membrane rest voltage. The function $f(\cdot)$ of the LIF neuron is defined as
\begin{equation}
    f\left(V^i_{t-1}, E^i_t\right)=V^i_{t-1}+\frac{1}{\tau}\left(-\left(V^i_{t-1}-V_{\text {r}}\right)+E^i_t\right),
\end{equation}
where $\tau$ is the membrane time constant. $\Theta(x)$ is the Heaviside step function, which is defined by $\Theta(x)=1$ for $x \geq 0$ and $\Theta(x)=0$ otherwise. Note that $V_0=V_{\text {r}}$.

Then the output of LIF layers regresses to the voltages of a smaller number of leaky integrate (LI) models. These voltages of LI neurons are prepared for the Q-value regression in reinforcement learning. The calculation of LI voltages is calculated by the LI neuron dynamics, which are expressed by
\begin{equation}
\begin{aligned}
    V^{j}_t = V^{j}_{t-1}+ \frac{1}{\tau}\left(-\left(V^{j}_{t-1}-V_{\text {r}}\right)+ \sum_{i=1}^{N} w^{\text{LIF}}_i \cdot S^{i}_t\right),
\end{aligned}
\end{equation}
where $V^j_t$ denote the $j$th membrane voltage after neural dynamics and the trigger of a spike at time-step $t$. $ w^{\text{LIF}}_i$ is the weight of the $i$th LIF neuron output. $N$ is the number of LIF neurons.

Next, the LI voltages are regressed to 2 Q-values through a fully connected layer, which are given by

\begin{equation}
\begin{aligned}
    Q^{\text{ON}}_t = \sum_{j=1}^{M}  \sigma(w^{\text{ON}}_j V^j_t + b^\text{ON}_j),
\end{aligned}
\end{equation}

\begin{equation}
\begin{aligned}
    Q^{\text{OFF}}_t = \sum_{j=1}^{M}  \sigma(w^{\text{OFF}}_j V^j_t + b^\text{OFF}_j),
\end{aligned}
\end{equation}
where $Q^{\text{ON}}_t$ and $Q^{\text{OFF}}_t$ denote the ON and OFF Q-values at time-step $t$, respectively. $w^{\text{ON}}_j$ and $w^{\text{OFF}}_j$ mean the weight of the $j$th LI neuron voltage for on and off value regression, respectively. $b^{\text{ON}}_j$ and $b^{\text{OFF}}_t$ are the bias of the $j$th LI neuron voltage for on and off trigger, respectively. $M$ is the number of LI neurons. $\sigma(\cdot)$ is the activation function.
Finally, the action $a_t$ is taken according the output Q-values for on and off trigger. That is,
\begin{equation}
    a_t =\Theta\left(Q^{\text{ON}}_t-Q^{\text{OFF}}_t\right),
\end{equation}
for which if $a_t = 1$, SEAN enables tracking or mapping estimation process. Conversely, if $a_t = 0$, SEAN let the estimator remain idle.

To train the network, the discrete Q learning strategy is adopted to update the weights and biases of SEAN including $w^{\text{LIF}}_i$, $w^{\text{ON}}_j$, $b^\text{ON}_j$, $w^{\text{OFF}}_j$ and $b^\text{OFF}_j$. Details of the back propagation method for this network may refer to \cite{DSQN}. The reward feedback is acquired from the estimator itself, which is introduced in the next subsection.


\subsection{Self Supervised Reward Construction of SEAN}

In order to update the SEAN weights, rewards are constructed for tracking and mapping trigger policy networks SEAN. While humans can adjust their hormonal states such as heart rate online during estimation, an online self-supervised strategy for reinforcement learning is adopted to ensure the lightweight, real-time, and generalizable nature of SEAN. Instead of relying on any external ground truth supervision for reward, the information that the estimator can output itself is used as feedback to adjust the network weights. This approach enables the network to adapt to arbitrary scenes and sensor configurations. Below, we present the reward design for mapping and tracking.

\subsubsection{Depth Generation and Fusion Reward for Mapping}
In order to define the reward for triggering mapping estimation, it is necessary to assess the ability of the mapping process to generate a valid reference depth frame for tracking. THE-SEAN uses the ESVO series as the baseline estimator. Within the ESVO framework, two distinct scenarios that influence the success of effective mapping and reference frame construction are identified: 1) the initialization of the depth map, and 2) the online updating of the reference depth frame.

During initialization, the effectiveness of mapping process depends on the number of valid depth points it initializes. A higher number of initialized depth points provides additional re-projection constraints for tracking and contribute to updating the reference depth frame, which results in more points for future updates. Consequently, the reward for mapping during initialization is defined as a function of the number of valid depth points initialized, which is given by
\begin{equation}
 R_{\text{init}}(t) = 
\begin{cases}
   N_{\text{SGM}}(t) - N_{\text{SGM}}(t-1), & a^{\text{init}}_t = 1, \\
   -\alpha, &  a^{\text{init}}_t = 0,
\end{cases} 
\end{equation}
where $R_{\text{init}}(t)$ is the reward for mapping initialization at time $t$. $N_{\text{SGM}}(t)$ is number of valid depth points generated by the modified semiglobal matching (SGM) algorithm\cite{SGM} during the initialization. $\alpha$ is a constant representing the punishment for the initialization delay. $a^{\text{init}}_t$ is the action taken for ESVO initialization. 


The effectiveness of the mapping update process is assessed by the number of points successfully fused during the depth fusion procedure. A larger number of fused points reflects a higher consistency between the current and previous frame mappings. Consequently, the reward for the mapping update is designed to be proportional to the number of fused points in the depth fusion process, as expressed by 
\begin{equation}
 R_{\text{map}}(t) = 
\begin{cases}
   N_{\text{BM}}(t) + \lambda_e N_{\text{e}}(t), & a^{\text{map}}_t = 1, \\
   \gamma^{\text{map}} N_{\text{BM}}(t_{\text{last}}) - \lambda_e N_{\text{e}}(t) + R^{\text{map}}_{\text{idle}}, &  a^{\text{map}}_t = 0,
\end{cases} 
\end{equation}
where $R_{\text{map}}(t)$ is the reward for mapping update at time $t$. $N_{\text{BM}}(t)$ is the number of valid depth fusion points generated from block matching (BM) algorithm. $N_{\text{e}}(t)$ is number of active events during a fixed interval (e.g. 30 ms). $\gamma^{\text{map}}$ is a decay factor representing the mapping information reduction during the idle interval. $t_\text{last}$ is the last time when a valid trigger is produced. $\lambda_e$ is a ratio to balance the scale between $N_{\text{BM}}(t)$ and $N_{\text{e}}(t)$. $a^{\text{map}}_t$ is the action taken for ESVO mapping estimation. 


\subsubsection{Fisher Information Matrix Trace Reward for Tracking}
In order to formulate the reward for triggering tracking estimation, it is essential to assess the information gain in pose estimation throughout the tracking process. Pose optimization is achieved through the reprojection of event-based representations to the referenced depth frame. During the optimization process, FIM is typically used to quantify the information gain from the observations. The trace of FIM serves as the metric for evaluating the information gain for pose tracking. 

In the optimization framework, FIM is equivalent to the Hessian matrix. To maintain the real-time performance, we approximate the trace of the Hessian using the Jacobian matrix and the optimized residuals. The information gain $I_{\text{track}}(t)$ of pose estimation at time $t$ is given by
\begin{equation}
    I_{\text{track}}(t) \approx \frac{\operatorname{Trace}(J_{\text{track}}(t)^\top J_\text{track}(t))} {\sum_{i=1}^M \text{res}^2_i(t)},
\end{equation}
where $J_{\text{track}}(t)$ is the Jacobian matrix corresponding to the the measurement model of event representation reprojection. Details on the Jacobian matrix computation may refer to \cite{ESVO}. $\text{res}_i(t)$ is the residual of each measurement after the optimization. $M$ is the total number of measurements.

Based on $I_{\text{track}}(t)$, the reward for triggering tracking process can be constructed, which is expressed by
\begin{equation}
 R_{\text{track}}(t) = 
\begin{cases}
   I_{\text{track}}(t) + \lambda_e N_{\text{e}}(t), & a^{\text{track}}_t = 1, \\
   \gamma^{\text{track}} I_{\text{track}}(t_{\text{last}}) - \lambda_e N_{\text{e}}(t) + R^{\text{track}}_{\text{idle}}, &  a^{\text{track}}_t = 0,
\end{cases} 
\end{equation}
where $R_{\text{track}}(t)$ is the reward for tracking pose estiamtion at time $t$. $\gamma^{\text{track}}$ is a decay factor representing the tracking information reduction during the idle interval. $t_\text{last}$ is the last time when a valid trigger is produced. $\lambda_e$ is a ratio to balance the scale between $I_{\text{track}}(t)$ and $N_{\text{e}}(t)$. $a^{\text{track}}_t$ is the action taken for ESVO tracking estimation. 


\subsection{Implementation of SEAN in Event-based Estimator}
\begin{table}[]
    \centering
    \caption{Parameter Settings of SEAN Implementation}
    \renewcommand\arraystretch{1.2}
    \belowrulesep=0pt
    \aboverulesep=0pt
\resizebox{0.35\textwidth}{!}{
\begin{tabular}{c|c|c}
\toprule
Component & Parameter & Value \\
\cmidrule{1-3} \multirow{3}{*}{SNN}
 & (LIF, LI, OUT) & (IN, 128, 2)  \\
 & Time Resolution & 0.001s  \\
 & Surrogate gradient & Sigmoid  \\
\cmidrule{1-3}  \multirow{5}{*}{Training} & Batch size & 32  \\
 & Replay buffer & 100/10  \\
 & Learning rate & 0.2 \\
 & Initial exploration rate & 0.8 \\
 & Exploration rate decay & 0.001 \\

\bottomrule
\end{tabular}}
\label{tab:param}
\end{table}

\begin{table*}[]
    \centering
    \caption{Comparison of Estimation Accuracy APE[cm] Between THE-SEAN and Latest Stereo Event Visual Odometry Algorithms.}
    \renewcommand\arraystretch{1.2}
    \belowrulesep=0pt
    \aboverulesep=0pt
\resizebox{0.67 \textwidth}{!}{
\begin{tabular}{cccccccc|cccc}
\toprule
\multirow{3}{*}{Dataset} & \multirow{3}{*}{Seq.} & \multicolumn{2}{c}{ESVO}  & \multicolumn{2}{c}{ES-PTAM} & \multicolumn{2}{c}{ESVO2 w/o IMU} & \multicolumn{2}{|c}{\textbf{TS-THE-SEAN}} & \multicolumn{2}{c}{\textbf{AA-THE-SEAN}} \\
\cmidrule(lr){3-4} \cmidrule(lr){5-6} \cmidrule(lr){7-8} \cmidrule(lr){9-10} \cmidrule(lr){11-12} 
&  & \multicolumn{2}{c}{APE} & \multicolumn{2}{c}{APE} & \multicolumn{2}{c}{APE} & \multicolumn{2}{|c}{APE} & \multicolumn{2}{c}{APE} \\

\cmidrule(lr){3-4} \cmidrule(lr){5-6} \cmidrule(lr){7-8} \cmidrule(lr){9-10} \cmidrule(lr){11-12} 
&  & RMS & STD & RMS & STD & RMS & STD & RMS & STD & RMS & STD \\
\cmidrule{1-12}
\multirow{4}{*}{RPG} & box  & 6.1 & 2.1 & 4.1 & 2.1 & 4.1 & 1.6 & 6.1 & 1.8 & \textbf{3.7} & \textbf{1.3} \\
 & monitor & 6.7 & 3.4 & 2.3 & 1.5 & 2.8 & 1.2 & 5.8 & 3.1 & \textbf{1.7} & \textbf{0.8} \\
 & bin & 4.1 & 1.4 & 2.6 & 0.9 & 2.5 & 0.8 & 3.4 & 1.1 &  \textbf{2.2} & \textbf{0.7} \\
 & desk & 3.4 & 1.2 & 2.8 & 1.5 & 2.5 & 1.0 & 2.7 & 0.9 & \textbf{2.2} & \textbf{0.9} \\

\cmidrule{1-12}
\multirow{3}{*}{MVSEC} & indoor1 &  15.9 & 5.6 & 15.0 & 6.3 & 9.6 & 4.9 & 10.5 & 4.5 & \textbf{8.6} & \textbf{4.4} \\
 & indoor2 &  16.6 & 5.3 & - & - & 14.7  & 7.1 & 13.0 & \textbf{5.3}  &  \textbf{11.5} & 5.5 \\
 & indoor3 &  10.2 & 4.9 & - & - & 9.0  & 4.8 &   8.2 & 3.9 & \textbf{7.4} & \textbf{3.5} \\

\cmidrule{1-12}
\multirow{5}{*}{DSEC} 

& city04a & 139.4  & 66.9 & 131.6 & 72.0 & 75.8 & 18.5 & 109.1 & 51.4  & \textbf{60.0} & \textbf{16.5} \\
& city04b & 42.9  & 20.9  & \textbf{29.0} & \textbf{13.2} & 63.7 & 24.1 & 42.0 & 17.7 & 60.4 & 21.7 \\
& city04c & 798.7 & 312.9 & 1184.4 & 588.8  & 571.1 & 241.5 & 730.4  & 290.3 & \textbf{549.2} & \textbf{241.1} \\
& city04d & 992.7 & 393.1 & 1053.9 & 349.7 & 615.5 & 266.7 & 833.5  & 347.8 & \textbf{509.1} & \textbf{226.7} \\
& city04e & 58.1  & 24.6 & 75.90 & 28.6 & 58.6 & 20.7 & 51.8 & 25.6 & \textbf{45.1} & \textbf{17.8} \\

\bottomrule
\multicolumn{4}{l}{\footnotesize{$^\ast$"-" means the lack of results of the algorithm.}}
\end{tabular}
}
\label{tab:accuracy comparison}
\end{table*}

THE-SEAN is implemented using the SpikingJelly\cite{SpikingJelly} framework designed for asynchronous event SNN processing. The detailed parameter settings are shown in Table \ref{tab:param}. 
The network structure is constructed according to Section IV-B. Our SEAN is configured with a input layer with LIF neurons for event spikes, a hidden layer with 128 LI neurons and an output layer of 2 neurons for Q-value regression. The time resolution of the network is set to 0.001 seconds, and a sigmoid function is employed for the surrogate gradient during training.
For the online self supervised weight update phase, the network is trained using a batch size of 32. The replay buffer is set to store 100/10 (100 for low resolution 346$\times$260 event camra and 10 for high resolution 640$\times$480) previous experiences for experience replay. The learning rate is initialized at 0.2, while the initial exploration rate is set to 0.8, decaying 0.001 over every training time-step to promote exploitation of learned policies as training progresses.
These parameter choices are designed to optimize the performance of SEAN while balancing computational efficiency and learning effectiveness.
\begin{table*}[]
    \centering
    \caption{Comparison of Triggering Rate Between THE-SEAN and Baseline Stereo Event Visual Odometry with TS/AA Representations.}
    \renewcommand\arraystretch{1.2}
    \belowrulesep=0pt
    \aboverulesep=0pt
\resizebox{0.88\textwidth}{!}{
\begin{tabular}{cccccccc|cccccc}
\toprule
\multirow{2}{*}{Dataset} & \multirow{2}{*}{Seq.} & \multicolumn{3}{c}{ESVO} & \multicolumn{3}{c}{\textbf{TS-THE-SEAN}} & \multicolumn{3}{|c}{ESVO2 w/o IMU} & \multicolumn{3}{c}{\textbf{AA-THE-SEAN}} \\
\cmidrule(lr){3-5} \cmidrule(lr){6-8} \cmidrule(lr){9-11} \cmidrule(lr){12-14} & & APE & TTR & MTR & APE & TTR & MTR & APE & TTR & MTR & APE & TTR & MTR\\
\cmidrule{1-14}
\multirow{4}{*}{RPG} & box & 6.1 & 99.5 & 19.6 & 6.1(== 0\%) & 80.0($\downarrow 20\%$) & 11.6($\downarrow 41\%$) & 4.1&  99.6&  19.8&  3.7($\downarrow 9.7\%$) &  84.1($\downarrow 16\%$)&  13.5($\downarrow 32\%$)\\
 & monitor & 6.7 & 99.7 & 19.1 & 5.8($\downarrow 13\%$) & 79.2($\downarrow 20\%$) & 10.0($\downarrow 48\%$) & 2.8&  99.7&  19.8&  1.7($\downarrow 39\%$) &  79.7($\downarrow 20\%$)&  15.4($\downarrow 22\%$)\\
 & bin & 4.1 & 99.5 & 19.0 & 3.4($\downarrow 17\%$)  & 78.3($\downarrow 20\%$) & 8.3($\downarrow 56\%$) & 2.5 &  100&  19.7&  2.2($\downarrow 12\%$) &  79.6($\downarrow 20\%$) &  16.3($\downarrow 17\%$)\\
 & desk & 3.4 & 100 & 19.8 & 2.7($\downarrow 21\%$) & 73.7($\downarrow 26\%$) & 19.1($\downarrow 3\%$) & 2.5&  99.5&  19.6&  2.2($\downarrow 12\%$) &  84.7($\downarrow 15\%$)&  15.2($\downarrow 23\%$)\\

\cmidrule{1-14}
\multirow{3}{*}{MVSEC} & indoor1 & 15.9 & 99.9 & 19.8 & 10.5($\downarrow 34\%$) & 56.5($\downarrow 43\%$) & 3.3($\downarrow 83\%$) & 9.6 & 100 & 19.9 & 8.6($\downarrow 10\%$) & 73.6($\downarrow 26\%$) & 8.3($\downarrow 59\%$) \\
 & indoor2 & 16.6 & 100 & 19.9 & 13.0($\downarrow 26\%$) & 65.6($\downarrow 35\%$) & 4.3($\downarrow 78\%$) & 14.7 & 100 & 20 & 11.5($\downarrow 22\%$)  & 74.6($\downarrow 25\%$) & 8.8($\downarrow 56\%$)  \\
 & indoor3 & 10.2 & 99.9 & 19.1 & 8.2($\downarrow 20\%$) & 56.4($\downarrow 43\%$) & 4.5($\downarrow 76\%$) & 9.0 & 100 & 19.9 & 7.4($\downarrow 18\%$) & 75.8($\downarrow 24\%$) & 7.9($\downarrow 60\%$) \\

\cmidrule{1-14}
\multirow{5}{*}{DSEC} 

& city04a & 139.4 & 94.6 & 19.9 & 109.1($\downarrow 22\%$)  & 82.1($\downarrow 13\%$) & 10.0($\downarrow 50\%$) & 75.8 & 100 & 19.9 & 60.0($\downarrow 21\%$) & 91.1($\downarrow 9\%$) & 12.1($\downarrow 39\%$) \\
& city04b & 42.9 & 94.5 & 19.6 & 42.0($\downarrow 2\%$) & 75.1($\downarrow 21\%$) & 10.6($\downarrow 46\%$) & 63.7 & 94.9 & 19.8 & 60.4($\downarrow 5\%$) & 81.4($\downarrow 14\%$) & 10.7($\downarrow 46\%$) \\
& city04c & 798.7 & 94.0 & 19.8 & 730.4($\downarrow 9\%$) & 84.1($\downarrow 11\%$) & 12.7($\downarrow 36\%$) & 571.1 & 94.0 & 19.9 & 549.2($\downarrow 4\%$) & 80.5($\downarrow 4\%$) & 12.6($\downarrow 37\%$) \\
& city04d & 992.7 & 93.9 & 19.9 & 833.5($\downarrow 16\%$) & 82.3($\downarrow 12\%$) & 8.3($\downarrow 58\%$) & 615.5 & 93.9 & 19.9 & 509($\downarrow 17.3\%$) & 95.7($\uparrow 2\%$) & 11.3($\downarrow 43\%$)\\
& city04e & 58.1 & 97.8 & 19.5 & 50.9 ($\downarrow 11\%$) & 87.3($\downarrow 12\%$) & 11.0($\downarrow 44\%$) & 58.6 & 97.7 & 19.8 & 45.1($\downarrow 23\%$) & 94.7($\downarrow 3\%$) & 14.9($\downarrow 25\%$) \\

\bottomrule
\multicolumn{4}{l}{\footnotesize{$^\ast$APE[cm] in RMS is listed in this table.}}
\end{tabular}}
\label{tab:MTR TTR}
\vspace{-0.5cm}
\end{table*}

\begin{figure}[]
\centering
\subfloat[Part of Seq. indoor1.]{\includegraphics[width=0.5 \linewidth]{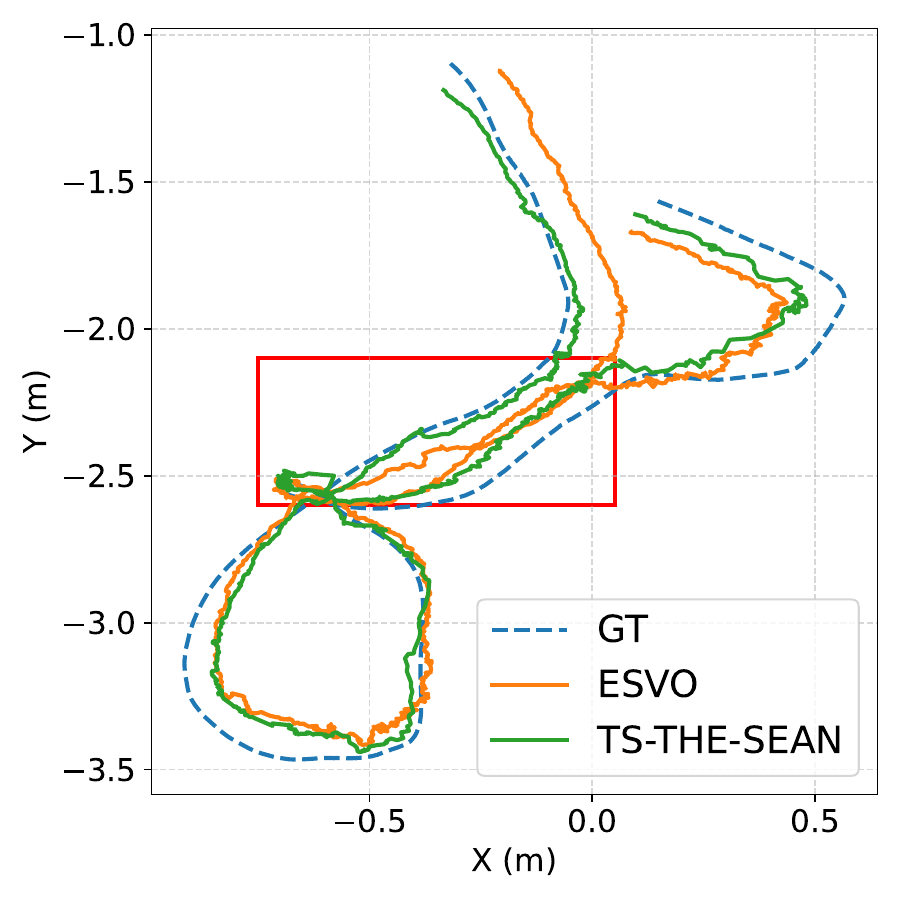}%
\label{fig_1_case}}
\subfloat[Part of Seq. indoor3.]{\includegraphics[width=0.5 \linewidth]{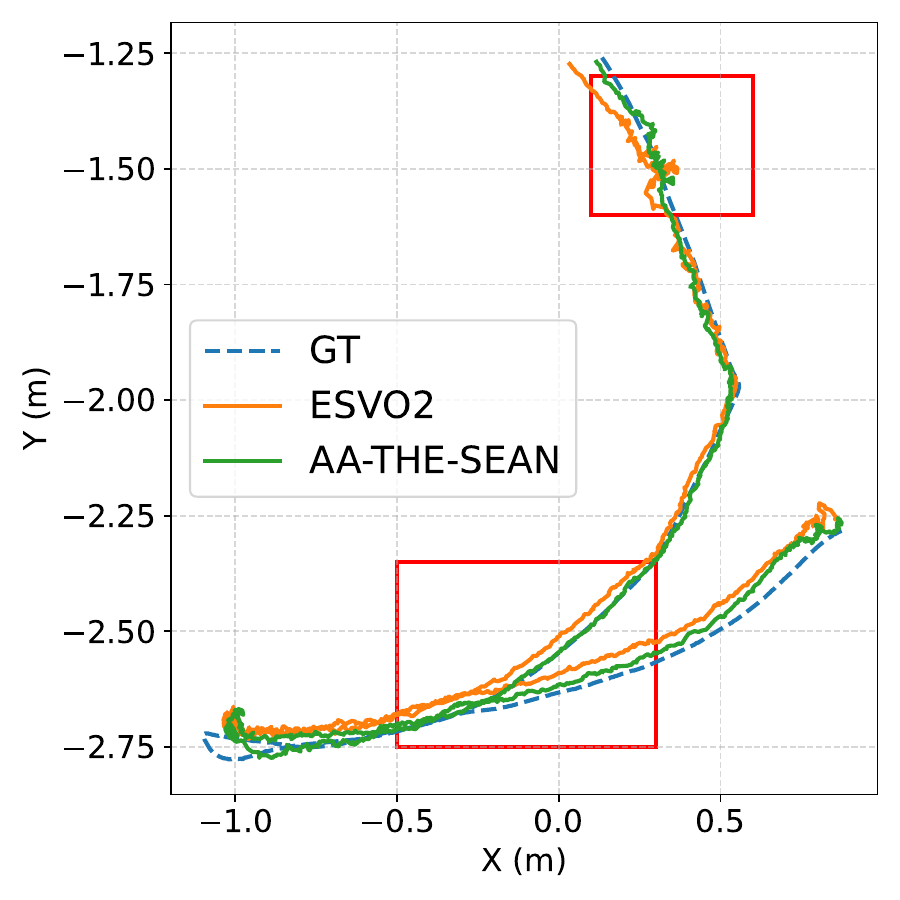}%
\label{fig_2_case}} \\
\subfloat[Agent velocity VS. MTR of TS-THE-SEAN on Seq. indoor1.]{\includegraphics[width=1 \linewidth]{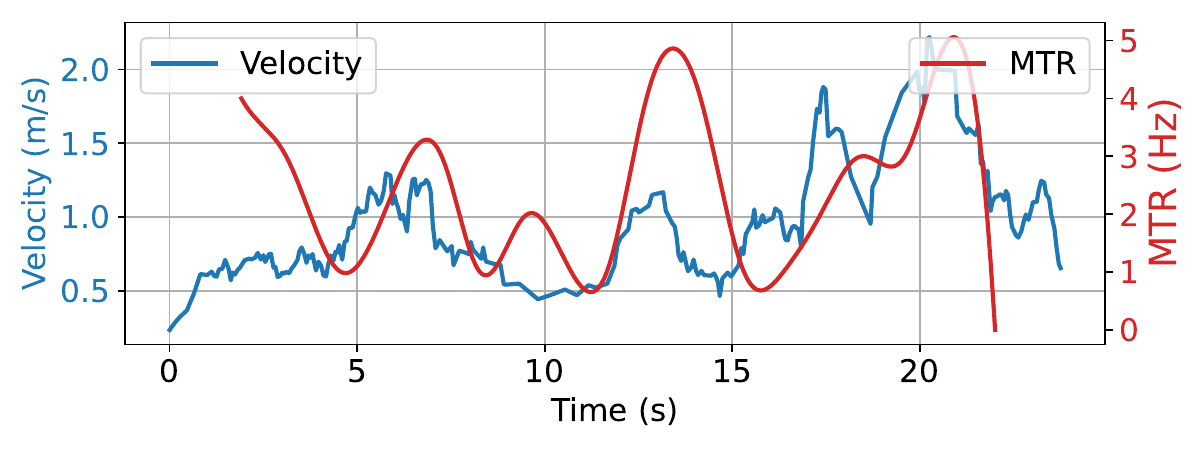}%
\label{fig_3_case}}
\caption{Trajectory comparison and MTR analysis. (a) and (b) illustrate part of the representative estimated trajectories by THE-SEAN and baselines on MVSEC. (c) shows the MTR variation of THE-SEAN corresponding to theagent velocity in sequence indoor1.}
\label{fig_MTR}
\vspace{-0.5cm}
\end{figure}

\section{EXPERIMENTS}
In this section, THE-SEAN is evaluated against some state-of-the-art event-based visual odometry algorithms. First, we introduce the experimental setup, evaluation metrics, and event-based datasets used for testing. Then, experiments conducted on popular open datasets is presented, focusing on two key aspects: 1) the overall estimation accuracy, and 2) the temporal computational efficiency of THE-SEAN.
\subsection{Experimental Setup}
\subsubsection{Evaluation Metrics}
To evaluate the overall estimation accuracy and the computational efficiency of the estimator, we develop the following three evaluation metrics:

\textit{Absolute Positioning Error (APE)}: APE is used to evaluate the overall estimation accuracy of the event-based estimator, with root mean square (RMS) and standard deviation (STD) metrics evaluating the absolute positioning accuracy and trajectory smoothness, respectively. APE is given in centimeters (cm) in this paper.

\textit{Tracking Triggering Rate (TTR)}: We define TTR to assess the energy consumption for tracking process of the estimator. TTR is the average triggering rate for tracking process for the event-based estimator; that is,
\begin{equation}
    \text{TTR} =  \frac{1}{N} \sum_{t=1}^{N} a^{\text{track}}_t,
\end{equation}
where $N$ is the length of the decision chain for tracking process. $a^{\text{track}}_t$ is the action taken for tracking estimation. This metric quantifies average tracking estimation times for the event-based estimator.

\textit{Mapping Triggering Rate (MTR)}: Similar to TTR, MTR is defined to assess the energy consumption for mapping process of the estimator. MTR is the average triggering rate for mapping process for the event-based estimator, which is given by
\begin{equation}
    \text{MTR} =  \frac{1}{N} \sum_{t=1}^{N} a^{\text{map}}_t,
\end{equation}
where $N$ is the length of the decision chain for mapping process. $a^{\text{map}}_t$ is the action taken for mapping estimation. This metric quantifies average mapping estimation times for the event-based estimator.

\subsubsection{Development of Experimental Datasets}
To demonstrate the effectiveness of THE-SEAN, experiments are conducted on public real-world datasets with various event resolutions and motion types.

\textit{RPG Dataset}\cite{RPG}: RPG is a hand-held stereo event camera dataset. The motion is relatively gentle and focus on one particular area. 

\textit{MVSEC Dataset}\cite{MVSEC}: MVSEC is a stereo event camera dataset collected by drones. The motion is relatively fierce and the variation of velocity is large. 

\textit{DSEC Dataset}\cite{DSEC}: DSEC is an autonomous driving dataset with stereo event cameras. The motion is very fierce, and the acceleration is large.

\subsubsection{Compared Algorithms}
THE-SEAN is compared against two main stereo-only event-based estimators, each utilizing different event representations, as follows:

\begin{itemize} 
\item ESVO: A classic real-time stereo-only event-based estimator using time surface (TS) event representation proposed by \warning{Zhou et al. in 2021}\cite{ESVO}.
\item ES-PTAM: A recent multi-camera event-based multi-view stereo (MC-EMVS) depth estimator designed for stereo-only event-based odometry \warning{introduced by Ghosh ea at. in 2024\cite{ES-PTAM}}.
\item ESVO2 w/o IMU: The latest stereo event-based estimator with TS event representation for tracking and adaptive accumulation (AA) for mapping \warning{presented by Niu et al. in 2024\cite{ESVO2}}. The original ESVO2 integrates IMU assistance, but for fair comparison, we modify it to a stereo-only event camera setup in this study. \end{itemize}

Note that the experiments focus on validating the temporally high-order strategies of the event-based estimator. Our SEAN is implemented with both the ESVO and ESVO2 w/o IMU estimators, referred to as TS-THE-SEAN and AA-THE-SEAN, to compare overall estimation accuracy and triggering rate. ES-PTAM is only included for comparison of overall estimation accuracy (APE) but cannot be directly compared for temporal computational efficiency (TTR or MTR) due to its high CPU demands, making it unsuitable for real-time implementation with SEAN.


\subsection{Experimental Results and Discussions}

\subsubsection{Overall Estimation Accuracy}

Table \ref{tab:accuracy comparison} compares the estimation accuracy of TS-THE-SEAN and AA-THE-SEAN with classic and state-of-the-art algorithms, including ESVO, ES-PTAM, and ESVO2 w/o IMU. On the RPG dataset, AA-THE-SEAN improves RMS by 18\% and STD by 19\% for APE on average. On the MVSEC indoor dataset, it improves RMS by 17\% and STD by 20\% on average. \warning{Fig. \ref{fig_MTR} (a) and (b) show part of the estimated trajectories by THE-SEAN and baselines.} On the DSEC indoor dataset, AA-THE-SEAN improves RMS by 11\% and STD by 8\% on average. These results demonstrate that AA-THE-SEAN achieves superior accuracy, along with smoother and more stable trajectories compared to existing methods.

\subsubsection{Triggering Rate Analysis}

Table \ref{tab:MTR TTR} compares the performance of TS-THE-SEAN and AA-THE-SEAN with ESVO and ESVO2 w/o IMU in terms of TTR and MTR across various datasets. TS-THE-SEAN improves TTR by 23\% and MTR by 51\% while AA-THE-SEAN improves TTR by 16\% and MTR by 38\% compared to their respective baselines across all testing sequences on average.
These results emphasize that TS-THE-SEAN and AA-THE-SEAN not only achieve higher estimation accuracy and stability, but also reduce the computational cost by triggering updates more efficiently. \warning{Fig. \ref{fig_MTR} (c) shows that THE-SEAN can dynamically adjust the estimation trigger decision to adapt to the agent motion.} The improvements in triggering rates are crucial for low-power systems, ensuring that the systems can operate with minimal computational overhead while maintaining robust performance in various scenarios.

\subsection{Ablation Study}
Table \ref{tab:ablation} presents the results of an ablation study evaluating the contributions of Mapping SEAN and Tracking SEAN in THE-SEAN across the RPG and DSEC datasets, measuring APE, TTR, and MTR.
The results of the ablation study confirms that both mapping SEAN and tracking SEAN significantly enhance THE-SEAN's performance, with the best balanced results achieved by combining both components, which optimize accuracy, efficiency, and computational cost.
\begin{table}[]
    \centering
    \caption{Ablation Study of THE-SEAN on RPG and DSEC Datasets.}
    \renewcommand\arraystretch{1.2}
    \belowrulesep=0pt
    \aboverulesep=0pt
\resizebox{0.4\textwidth}{!}{
\begin{tabular}{c|c|c|c|c|c}
\toprule
Dataset & Mapping SEAN  & Tracking SEAN & APE & TTR & MTR\\
\cmidrule{1-6} \multirow{4}{*}{RPG}
  & $\times$ & $\times$ & 2.98 & 99.7 & 19.7 \\
  & \checkmark & $\times$ & 2.55 & 98.86 & 15.49 \\
 & $\times$ & \checkmark & 2.68 & 79.63 & 19.89\\
 & \checkmark & \checkmark  & 2.45 & 82.03 & 15.1 \\

 \cmidrule{1-6} \multirow{4}{*}{DSEC}
  & $\times$ & $\times$ & 203.4 & 96.6 & 19.9 \\
 & \checkmark & $\times$ & 167.7 & 96.24 & 12.46  \\
 & $\times$ & \checkmark & 186.7 &  87.16 & 18.95 \\
 & \checkmark & \checkmark & 168.6 & 90.74 & 12.24\\

\bottomrule
\end{tabular}}
\label{tab:ablation}

\end{table}

\vspace{-0.5cm}
\subsection{Computational Cost Analysis}
Table \ref{tab:cost} presents the computational cost analysis of the various modules in AA-THE-SEAN, measured in the number of operations (OPs) per module. The tracking and mapping modules of the baseline ESVO2 w/o IMU, crucial for state estimation, necessitate 1800M (39.4\%) and 2600M (56.9\%) OPs, respectively, highlighting their resource-intensive nature. Conversely, SEAN, which utilizes lightweight spiking networks, only requires 69M (1.5\%) for per back propagation. Moreover, SEAN can save about 16\% tracking triggering times and 38\% mapping triggering times. This results in a substantial reduction in computational cost while maintaining high estimation performance. The SEAN module’s low OPs, combined with the inherent SNN low-power nature, makes it well-suited for real-time, event-based state estimation in dynamic environments. 

\begin{table}[]
    \centering
    \caption{Number of Operations and Proportions for each Module in AA-THE-SEAN.}
    \renewcommand\arraystretch{1.2}
    \belowrulesep=0pt
    \aboverulesep=0pt
\resizebox{0.45\textwidth}{!}{
\begin{tabular}{c|cccc|c|c}
\toprule
\multirow{2}{*}{Module} &  \multicolumn{4}{c|}{ESVO2 w/o IMU} & \textbf{SEAN} & AA-THE-SEAN \\
\cmidrule(lr){2-5}   \cmidrule(lr){6-6}  \cmidrule(lr){7-7}  
& TS  & AA & Tracking & Mapping & \textbf{BP} & SUM \\
\cmidrule{1-7}
OPs & 39M & 63M & 1800M & 2600M & \textbf{69M} & 4571M\\
Proportion & 0.8\% & 1.4\% & 39.4\% & 56.9\% & \textbf{1.5\%} & 100\%\\

\bottomrule
\end{tabular}}
\label{tab:cost}
\vspace{-0.5cm}
\end{table}
\section{CONCLUSIONS}
In this paper, we introduce THE-SEAN, a temporally high-order event-based visual odometry system utilizing self-supervised spiking event accumulation networks. Inspired by biological mechanisms regulating heart rate, THE-SEAN dynamically adjusts its estimation trigger decision policy based on changes in motion and the environment. Experimental results demonstrate that THE-SEAN not only enhances estimation accuracy and smoothness but also significantly improves triggering efficiency compared to state-of-the-art methods. Future work will focus on the neuromorphic hardware implementation of THE-SEAN and its integration with synchronous sensors, such as IMUs and traditional cameras, to further optimize the temporally high-order system.

\addtolength{\textheight}{-12cm}   





\normalem
\bibliographystyle{IEEEtran}
\bibliography{ref}

\end{document}